\documentclass[runningheads]{llncs}
\usepackage{graphicx}
\usepackage{amsmath,amssymb} 
\usepackage{color}
\usepackage{url}
\usepackage{algorithmic}
\usepackage{listings}
\usepackage{epstopdf}
\begin{document}
	\pagestyle{headings}
	\mainmatter

	\def\ACCV{}
	\title{A Coarse-to-Fine Indoor Layout Estimation (CFILE) Method} 
	
	\titlerunning{}
	\authorrunning{}

	\author{Yuzhuo Ren, Chen Chen, Shangwen Li, and C.-C. Jay Kuo}
	\institute{}
	
	\maketitle
	
	\begin{abstract}
		
		The task of estimating the spatial layout of cluttered indoor scenes
		from a single RGB image is addressed in this work.  Existing solutions
		to this problems largely rely on hand-craft features and vanishing
		lines, and they often fail in highly cluttered indoor rooms. The
		proposed coarse-to-fine indoor layout estimation (CFILE) method consists
		of two stages: 1) coarse layout estimation; and 2) fine layout
		localization.  In the first stage, we adopt a fully convolutional neural
		network (FCN) to obtain a coarse-scale room layout estimate that is
		close to the ground truth globally. The proposed FCN considers combines
		the layout contour property and the surface property so as to provide a
		robust estimate in the presence of cluttered objects.  In the second
		stage, we formulate an optimization framework that enforces several
		constraints such as layout contour straightness, surface smoothness and
		geometric constraints for layout detail refinement. Our proposed system
		offers the state-of-the-art performance on two commonly used benchmark
		datasets. 
		
	\end{abstract}
	
	\section{Introduction}\label{sec:introduction}
	
	The task of spatial layout estimation of indoor scenes is to locate the
	boundaries of the floor, walls and the ceiling. It is equivalent to the
	problem of semantic surface labeling. The segmented boundaries and
	surfaces are valuable for a wide range of computer vision applications
	such as indoor navigation \cite{karsch2011rendering}, object detection
	\cite{hedau2010thinking} and augmented reality \cite{karsch2011rendering,liu2015rent3d,xiao2014reconstructing,martin20143d}. Estimating
	the room layout from a single RGB image is a challenging task. This is
	especially true in highly cluttered rooms since the ground and wall
	boundaries are often occluded by various objects. Besides, indoor scene
	images can be shot at different viewpoints with large intra-class
	variation. As a result, high-level reasoning is often required to avoid
	confusion and uncertainty. For example, the global room model and its
	associated geometric reasoning can be exploited for this purpose. Some
	researchers approach this layout problem by adding the depth information
	\cite{zhang2013estimating,guo2015predicting}. 
	
	The indoor room layout estimation problem has been actively studied in
	recent years. Hedau {\em et al.} \cite{hedau2009recovering} formulated
	it as a structured learning problem. It first generates hundreds of
	layout proposals based on inference from vanishing lines. Then, it uses
	the line membership features and the geometric context features to rank
	the obtained proposals and chooses the one with the highest score as the
	desired final result. 
	
	\begin{figure}
		\centering
		\includegraphics[width=120mm]{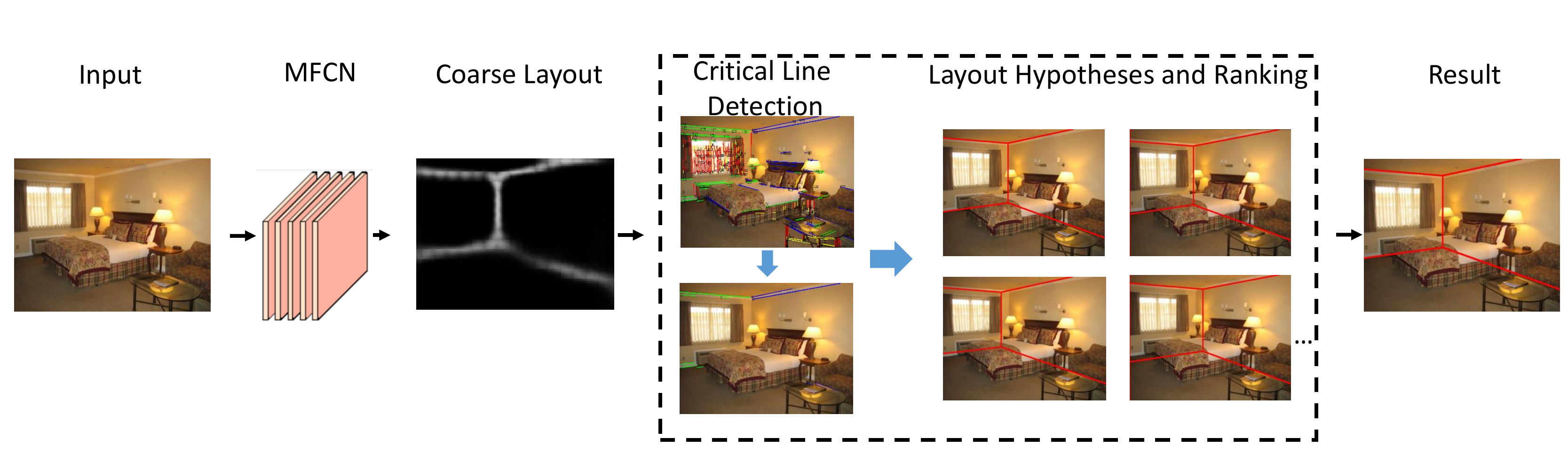} 
		\caption{The pipeline of the proposed coarse-to-fine indoor layout
			estimation (CFILE) method. For an input indoor image, a coarse layout
			estimate that contains large surfaces and their boundaries is obtained
			by a multi-task fully convolutional neural network (MFCN) in the first stage. Then, occluded lines and missing lines are
			filled in and possible layout choices are ranked according to a
			pre-defined score function in the second stage. The one with the highest
			score is chosen to the final output.} \label{fig:pipline}
	\end{figure}
	
	In this work, we propose a coarse-to-fine indoor layout estimation
	(CFILE) method. Its pipeline is shown in Fig. \ref{fig:pipline}.  The
	system uses an RGB image as its input and provides a box layout as its
	output. The CFILE method consists of two stages: 1) coarse layout
	estimation; and 2) fine layout localization.  In the first stage, we
	adopt a multi-task fully convolutional neural network (MFCN) \cite{dai2015instance} to
	obtain a coarse-scale room layout estimate. This is motivated by the
	strength of the FCN in semantic segmentation \cite{long2015fully} and contour detection
	\cite{xie2015holistically}. The FCN has a strong discriminant power in
	handling a large variety of indoor scenes using the surface property and
	the layout contour property. It can provide a robust estimate in the
	presence of cluttered objects, which is close to the ground truth
	globally.  In the second stage, being motivated by structured learning,
	we formulate an optimization framework that enforces several constraints
	such as layout contour straightness, surface smoothness and geometric
	constraints for layout detail refinement. 
	
	It is worthwhile to emphasize that the spatial layout estimation problem
	is different from semantic object segmentation problem in two aspects.
	First, the spatial layout problem targets at the labeling of semantic
	surface of an indoor room rather than objects in the room.  Second, we
	have to label occluded surfaces while semantic segmentation does not
	deal with the occlusion problem at all. It is also different from the
	contour detection problem since occluded layout contours have to be
	detected. 
	
	The major contributions of this work are three folds.  First, we use the
	FCN to learn the labeling of main surfaces and key contours jointly,
	which are critical to robust spatial layout of an indoor scene. The FCN
	training is elaborated. It is shown that the course-scale layout
	estimate obtained by the FCN is robust and close to ground truth.
	Second, we formulate an optimization framework that enforces three
	contraints (i.e.  surface smoothness, contour straightness and proper
	geometrical structure) to refine the coarse-scale layout estimate.
	Third, we conduct extensive performance evaluation by comparing the
	proposed CFILE method and several benchmarking methods on the dataset of
	Hedau {\em et al.} \cite{hedau2009recovering}, the LSUN validation
	dataset \cite{maldives}. It is shown by experimental results that the
	proposed CFILE method offers the state-of-the-art performance. It
	outperforms the second best method by 1.16\% and 1.32\% in Hedau's
	dataset and the LSUN dataset, respectively. 
	
	The rest of this paper is organized as follows. Related previous work is
	reviewed in Sec. \ref{RW}. The proposed CFILE method is described in
	detail in Sec. \ref{method}. Experimental results are shown in Sec.
	\ref{result}. Concluding remarks are drawn in Sec. \ref{conclusion}. 
	
	\section{Related Work}\label{RW}
	
	{\bf Structured Learning.} The structured learning methodology
	\cite{nowozin2011structured} has been widely used in the context of
	indoor room layout estimation.  It targets at learning the structure of
	an environment in the presence of imperfect low-level features.  It
	consists of two stages \cite{nowozin2011structured}.  First, a set of
	structure hypotheses are generated. Second, a score function is defined
	to evaluate the structure in hypotheses set. The first stage is guided
	by low level features such as vanishing lines under the Manhattan
	assumption.  The number of layout hypotheses in the first stage is
	usually large while most of them are of low accuracy due to the presence
	of clutters. If the quality of hypotheses is low in the first stage,
	there is no easy way to fix it in the second stage. In the second stage
	of layout ranking, the score function contains various features such as
	the line membership \cite{hedau2009recovering,mallya2015learning}, the
	geometric context \cite{hedau2009recovering,mallya2015learning}, the
	object location \cite{gupta2010estimating}, etc. The score function
	cannot handle objects well since they overlap with more than one surfaces
	(e.g., between the floor and walls). The occluding objects in turn make
	the surface appearance quite similar along their boundaries. 
	
	{\bf Classical Methods for Indoor Layout Estimation.} Research on indoor
	room layout estimation has been active in recent years. Hedau {\em et
		al.} \cite{hedau2009recovering} formulated it as a structured learning
	problem.  There are many follow-up efforts after this milestone work.
	They focus on either developing new criteria to reject invalid layout
	hypotheses or introducing new features to improve the score function in
	layout ranking. 
	
	Different hypothesis evaluation methods were considered in 
	\cite{guo2015predicting,hedau2009recovering,gupta2010estimating,schwing2013box,zhao2013scene,pero2012bayesian,ramalingam2013manhattan}. Hedau
	{\em et al.} \cite{hedau2009recovering} reduced noisy lines by removing
	clutters first.  Specifically, they used the line membership together
	with semantic labeling to evaluate hypotheses.  Gupta {\em et al.}
	\cite{gupta2010estimating} proposed an orientation map that labels three
	orthogonal surface directions based on line segments and, then, used the
	orientation map to re-evaluate layout proposals. Besides, they detected
	objects and fit them into 3D boxes.  Since an object cannot penetrate
	the wall, they used the box location as a constraint to reject invalid
	layout proposals.  The work in
	\cite{hedau2010thinking,wang2013discriminative} attempted to model
	objects and spatial layout simultaneously. Hedau {\em et al.}
	\cite{hedau2012recovering} improved their earlier work in
	\cite{hedau2010thinking,hedau2009recovering} by localizing the box more
	precisely using several cues such as edge- and corner-based features.
	Ramalingam {\em et al.} \cite{ramalingam2013manhattan} proposed an
	algorithm to detect Manhattan Junctions and selected the best layout by
	optimizing a conditional random field whose corners are well aligned
	with pre-detected Manhattan Junctions. Pero {\em et al.}
	\cite{pero2012bayesian} integrated the camera model, an enclosing room
	box, frames (windows, doors, pictures), and objects (beds, tables,
	couches, cabinets) to generate layout hypotheses.  Lampert {\em et al.}
	\cite{lampert2009efficient} improved objects detection by maximizing a
	score function through the branch and bound algorithm. 
	
	{\bf 3D- and Video-based Indoor Layout Estimation.} Zhao and Zhu
	\cite{zhao2013scene} exploited the location information and 3D spatial
	rules to obtain as many 3D boxes as possible. For example, if a bed is
	detected, the algorithm will search its neighborhood to look for a side
	table.  Then, they rejected impossible layout hypothesis.  Choi {\em et
		al.} \cite{choi2013understanding} trained several 3D scene graph models
	to learn the relation among the scene type, the object type, the object
	location and layout jointly. Guo {\em et al.} \cite{guo2015predicting}
	recovered 3D model from a single RGBD image by transferring the exemplar
	layout in the training set to the test image. Fidler {\em et al.} \cite{fidler20123d} and Xiang {\em et al.} \cite{xiang2012estimating} represented objects
	by a deformable 3D cuboid model for improved object detection and then
	used in layout estimation.
	Fouhey {\em et al.} \cite{fouhey2014people} exploited human action and
	location in time-lapse video to infer functional room geometry. 
	
	{\bf CNN- and FCN-based Indoor Layout Estimation.} The convolution
	neural network (CNN) has a great impact on various computer vision
	research topics, such as object detection, scene classification,
	semantic segmentation, etc. Mallya and Lazebnik
	\cite{mallya2015learning} used the FCN to learn the informative edge
	from an RGB image to provide a rough layout.  The FCN shares features in
	convolution layers and optimize edges detection and geometric context labeling \cite{hedau2009recovering,hoiem2005geometric,hoiem2007recovering}
	jointly. The learned contours are used as a new feature in sampling
	vanishing lines for layout hypotheses generation. Dasgupta {\em et al.} \cite{Dasgupta2016Robust} used the FCN to learn
	semantic surface labels. Instead of learning edges, their solution
	adopted the heat map of semantic surfaces obtained by the FCN as the
	belief map and optimized it furthermore by vanishing lines. Generally speaking, a
	good layout should satisfy several constraints such as boundary
	straightness, surface smoothness and proper geometrical structure.
	However, the CNN is weak in imposing spatial constraints and performing
	spatial inference.  As a result, an inference model was appended in both
	\cite{mallya2015learning} and \cite{Dasgupta2016Robust} to refine the
	layout result obtained by CNN. 
	
	\section{Coarse-to-Fine Indoor Layout Estimation (CFILE)}\label{method}
	
	\subsection{System Overview}\label{overview}
	
\begin{figure}
		\centering
		\includegraphics[width=120mm]{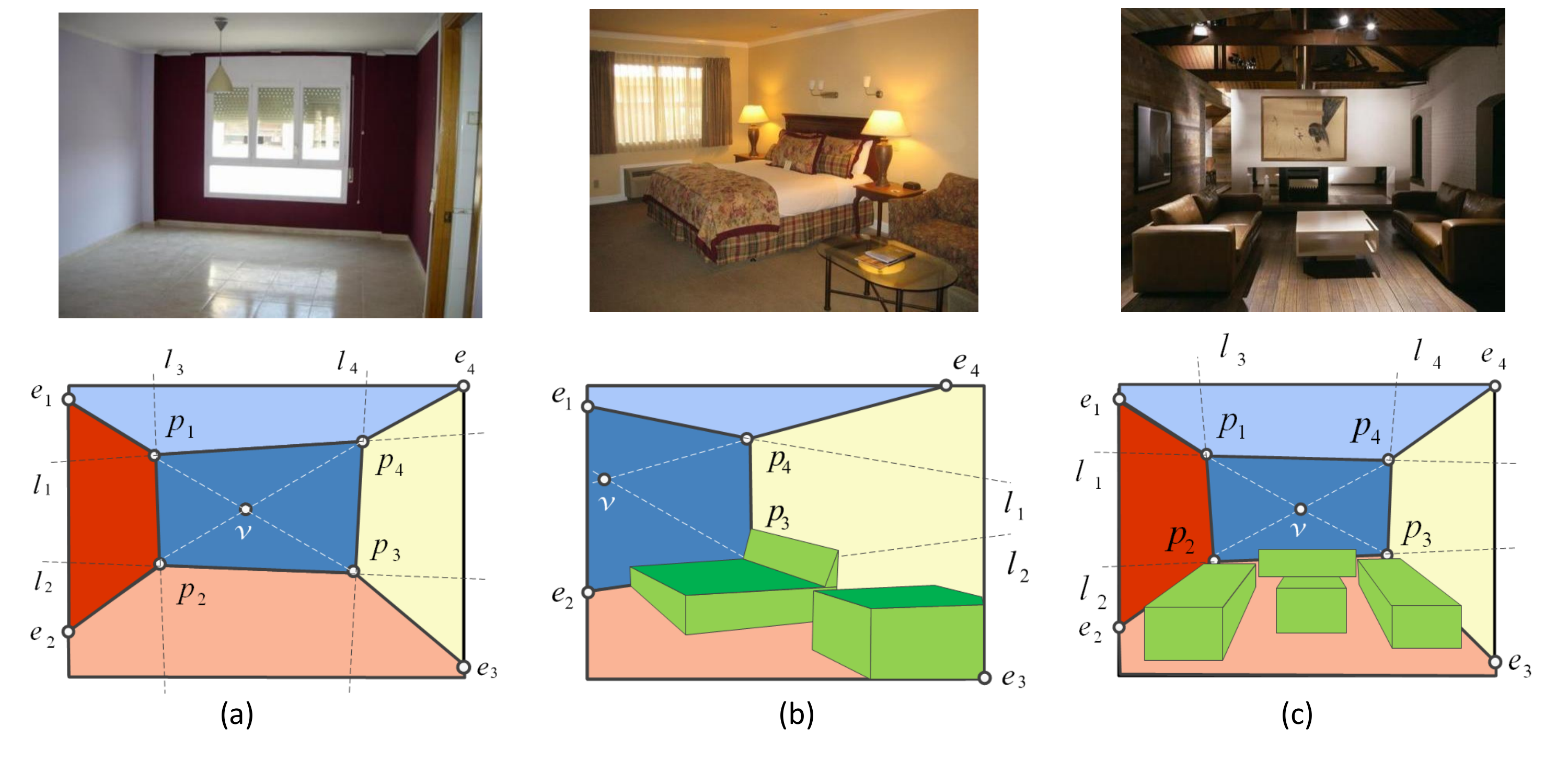} 
		\caption{Illustration of a layout model $\text{Layout}=(l_{1},l_{2},l_{3},l_{4},v)$
			that is parameterized by four lines and a vanishing point: (a) an easy
			setting where all five surfaces are present; (b) a setting where some
			surfaces are outside the image; (c) a setting where key boundaries are
			occluded.} \label{fig:LayoutModel}
	\end{figure}

	Most research on indoor layout estimation \cite{guo2015predicting,hedau2009recovering,gupta2010estimating,schwing2013box,zhao2013scene,pero2012bayesian,ramalingam2013manhattan} is based on the
	``Manhattan World" assumption. That is, a room contains three orthogonal
	directions indicated by three groups of vanishing lines. Hedau {\em et al.}
	\cite{hedau2009recovering} presented a layout model based on 4 rays
	and a vanishing point. The model can written as
	\begin{equation}\label{equ:layout_parameter} 
	\text{Layout}=(l_{1},l_{2},l_{3},l_{4},v),
	\end{equation}
	where $l_{i}$ is the $i^{th}$ line and $v$ is the vanishing point.  If
	$(l_{1},l_{2},l_{3},l_{4},v)$ can be easily detected without any
	ambiguity, the layout problem is straightforward.  One example is given
	in Fig. \ref{fig:LayoutModel} (a), where five surfaces are visible in
	the image without occlusion.
	
	However, more challenging cases exist. Vertices $p_{i}$ and $e_{i}$ in
	Fig. \ref{fig:LayoutModel} (a) may lie outside the image. One
	example is shown in Fig.  \ref{fig:LayoutModel} (b). Furthermore,
	vertices $p_{2}$ and $p_{3}$ are floor corners and they are likely be
	occluded by objects. Furthermore, line $l_{2}$ may be entirely or
	partially occluded as shown in Fig. \ref{fig:LayoutModel} (c). Lines
	$l_{3}$ and $l_{4}$ are wall boundaries, and they can be partially
	occluded but not fully occluded. Line $l_{1}$ is the ceiling boundary
	which is likely to be visible. 
	
	The proposed CFILE system consists of two stages as illustrated in Fig.
	\ref{fig:pipline}. In the fist stage, we propose a multi-task fully
	convolutional neural network (MFCN) to offer a coarse yet robust layout
	estimation.  Since the CNN is weak in imposing spatial smoothness and
	conducting geometric reasoning, it cannot provide a fine-scale layout
	result.  In the second stage, we first use the coarse layout from MFCN
	as the guidance to detect a set of critical lines. Then, we generate a
	small set of high quality layout hypotheses based on these critical
	lines. Finally, we define a score function to select the best layout as
	the desired output. Detailed tasks in these two stages are elaborated
	below. 
	
	\subsection{Coarse Layout Estimation via MFCN}\label{MFCN} 
	
\begin{figure}[htb]
		\centering
		\includegraphics[width=120mm]{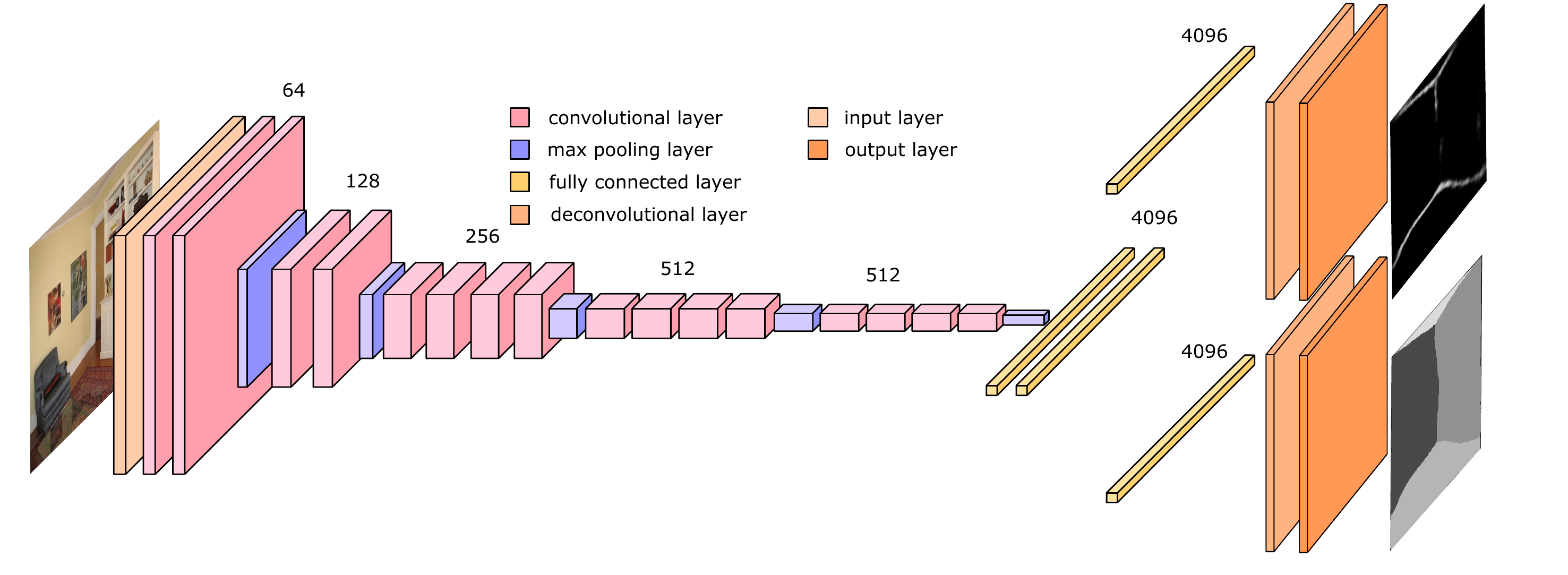} 
		\caption{Illustration of the FCN-VGG16 with two output branches. We use
			one branch for the coarse layout learning and the other branch for
			semantic surface learning. The input image size is re-sized to
			$404\times404$ to match the receptive field size of the filter at the
			fully connection layer.} \label{fig:CNN_flowchart}
	\end{figure}

	We adopt a multi-task fully convolutional neural network (MFCN)
	\cite{long2015fully,mallya2015learning,dai2015instance} to learn the
	coarse layout of indoor scenes. The MFCN \cite{dai2015instance} shares
	features in the convolutional layers with those in the fully connected
	layers and builds different branches for multi-task learning. The total
	loss of the MFCN is the sum of losses of different tasks.  The proposed
	two-task network structure is shown in Fig.  \ref{fig:CNN_flowchart}. We
	use the VGG-16 architecture for fully convolutional layers and train the
	MFCN for two tasks jointly, i.e. one for layout learning while the other
	for semantic surface learning (including the floor, left-, right-,
	center-walls and the ceiling). Our work is different from that
	in \cite{mallya2015learning}, where layout is trained together with geometric context labels \cite{hoiem2005geometric,hoiem2007recovering} which contains object labels. Here, we train the layout and semantic
	surface labels jointly. By removing objects from the concern, the
	boundaries of semantic surfaces and layout contours can be matched even
	in occluded regions, leading to a clearer layout. As compared to the
	work in \cite{Dasgupta2016Robust}, which adopts the fully convolutional
	neural network to learn semantic surfaces with a single task network,
	our network has two branches, and their learned results can help each
	other. 
	
	The receptive field of the filter at the fully connected layer of the
	FCN-VGG16 is $404\times404$, which is independent of the input image
	size \cite{long2015fully,xu2015multi}. Xu {\em et al.}
	\cite{xu2015multi} attempted to vary the FCN training image size so
	as to capture different level of details in image content. If the input
	image size is larger than the receptive field size, the filter of the
	fully connected layer looks at a part of the image.  If the input image
	size is smaller than the receptive field size, it is padded with zeros
	and spatial resolution is lost in this case.  The layout describes the
	whole image's global structure. We resize the input image to
	$404\times404$ so that the filter examines the whole image.    
 
\subsection{Layout Refinement}

There are two steps in structured learning: 1) to generate a hypotheses
set; and 2) to define a score function and search a structure in the
hypotheses set that maximizes the score function. We attempt to improve
in both areas. 

Given an input image \textbf{I} of size $w \times h \times 3$, the output of the
coarse layout from the proposed MFCN in Fig. \ref{fig:CNN_flowchart} is
a probability function in form of
\begin{equation}\label{equ:probability}
\textbf{P}^{(k)} = Pr(\textbf{L}_{ij}=k|\textbf{I}),  \quad \forall k\in 
\lbrace 0,1 \rbrace, \; i\in \lbrack 1,...,h \rbrack, \; j\in \lbrack 1,...,w\rbrack,
\end{equation}
where $\textbf{L}$ is an image of size $w \times h$ that maps each pixel
in the original image, $\textbf{I}_{ij}$, to a label in the output image
$\textbf{L}_{ij} \in \lbrace \text{0,1}\rbrace$, where 0 denotes
a background pixel and 1 denotes a layout pixel. One way to
estimate the final layout from the MFCN output is to select the label 
with the highest score; namely,
\begin{equation}\label{equ:L_estimate}
\hat{\textbf{L}}_{ij} = \operatornamewithlimits{argmax}\limits_{k} 
\textbf{P}_{ij}^{(k)} \quad \forall i\in \lbrack 1,...,h \rbrack, \;
j\in \lbrack 1,...,w\rbrack.
\end{equation}

It is worthwhile to point out that $\hat{\textbf{L}}_{ij}$ generated
from the MFCN output is noisy for two reasons.  First, the contour from
the MFCN is thick and not straight since the convolution operation and
the pooling operation lose the spatial resolution gradually along
stages. Second, the occluded floor boundary (e.g., the $l_2$ line in
Fig. \ref{fig:LayoutModel}) is more difficult to detect since it is less
visible than other contours (e.g., the $l_1$, $l_3$ and $l_4$ lines in
Fig. \ref{fig:LayoutModel}).  We need to address these two challenges in
defining a score function. 

The optimal solution for Eq. (\ref{equ:L_estimate}) is difficult to get
directly.  Instead, we first generate layout hypotheses that are close
to the global optimal layout, denoted by $\textbf{L}^{*}$, in the layout
refinement algorithm. Then, we define a novel score function to rank
layout hypotheses and select the one with the highest score as the final
result. 
 
\subsubsection{Generation of High-Quality Layout Hypotheses}

Our objective is to find a set of layout hypotheses that contains fewer
yet more robust proposals in the presence of occluders.  Then, the best
layout with the smallest error can be selected. 
 
\begin{figure}
\centering
\includegraphics[width=120mm]{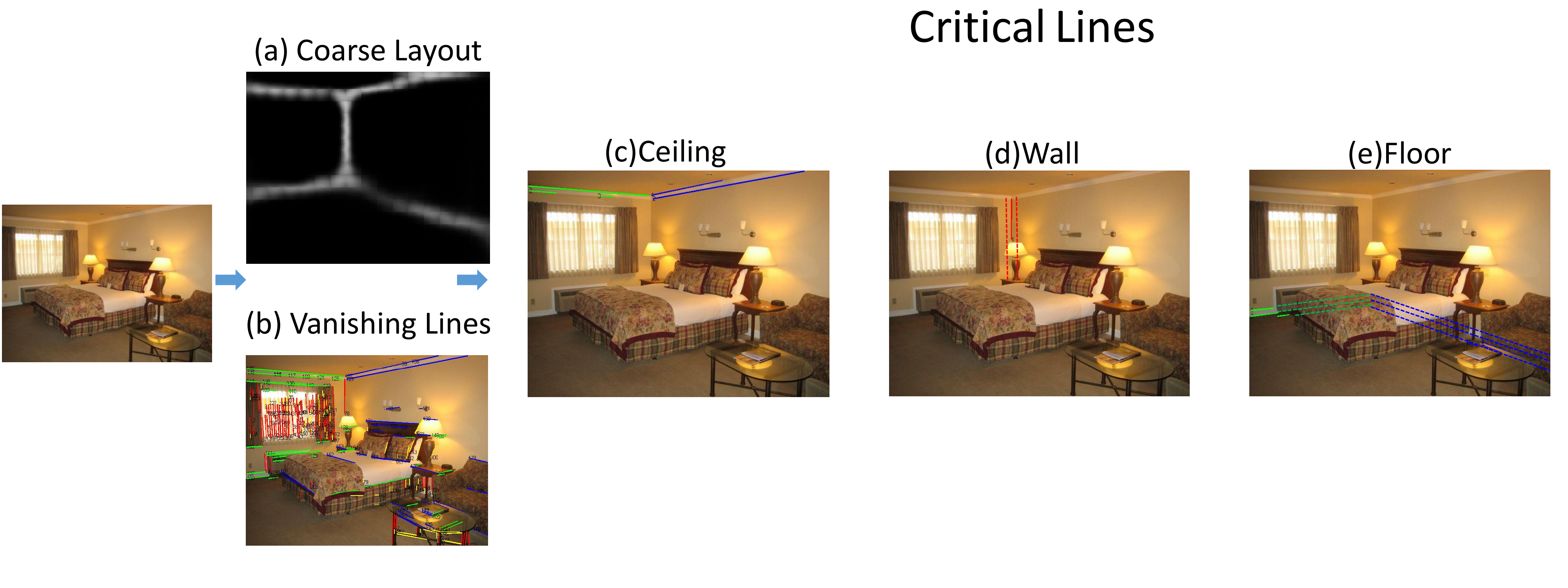} 
\caption{Illustration of critical lines detection for better layout
hypothesis generation. For a given input image, the coarse layout offers
a mask that guides vanishing lines selection and critical lines
inference. The solid lines indicate detected vanishing lines $C$. The dashed wall lines
indicate those wall lines that are not detected but inferred inside mask
$C$ from ceiling corners. The dashed floor lines indicate those floor lines that are not
detected but inferred inside mask $C$.}\label{fig:critical_line_floor}
\end{figure}

\textbf{Vanishing Line Sampling}. We first threshold the layout contour
obtained by the MFCN, convert it into a binary mask, and dilate it by 4
pixels to get a binary mask image denoted by $C$. Then, we apply the
vanishing lines detection algorithm \cite{gupta2010estimating} to the
original image and select those inside the binary mask as critical lines
$l_{i(\text{original})}$, shown in solid lines in Fig. \ref{fig:critical_line_floor} (c) (d) (e) for ceiling, wall and floor separately. Candidate vanishing point $v$ is generated by grid search around the initial
$v$ from \cite{gupta2010estimating}. 

\textbf{Handling Undetected Lines}. There is case when no vanishing lines are detected inside $C$ because of low contrast, such as wall boundaries, $l_{3}$(or $l_{4}$). If ceiling corners are available, $l_{3}$(or $l_{4}$) are filled in by connecting ceiling corners and vertical vanishing point. If ceiling corners do not present in the image, the missing $l_{3}$(or $l_{4}$) is estimated by logistic regression use the layout points in $\textbf{L}$. 

\textbf{Handling Occluded Lines}. As discussed earlier, the floor line,
$l_{2}$, can be entirely or partially occluded. One illustrative example
is shown in Fig. \ref{fig:critical_line_floor} where $l_{2}$ is
partially occluded. If $l_{2}$ is partially occluded, the occluded part
of $l_{2}$ can be recovered by line extension. For entirely occluded
$l_{2}$, if we simply search lines inside $C$
or uniformly sample lines \cite{mallya2015learning}, the layout proposal is not going to be
accurate as the occluded boundary line cannot be recovered. Instead, we
automatically fill in occluded lines based on geometric rule. If $p_{2}$ (or $p_{3}$) is
detectable by connecting detected $l_{3}$ (or $l_{4}$) to $e_{2}v$(or
$e_{3}v$), $l_{2}$ is computed as the line passing through the available
$p_{2}$ or $p_{3}$ and the vanishing point $l_{2}$ associated with. If
neither $p_{2}$ nor $p_{3}$ is detectable, $l_{2}$ is estimated by logistic regression use the layout points in $\textbf{L}$.

In summary, the final $l_{\text{critial}}$ used in generating layout hypotheses
is the union of three parts as given below:
\begin{equation}\label{equ:L}
l_{\text{critical}} = l_{i(\text{original})} \cup l_{i(\text{occluded})} \cup
l_{i(\text{undetected})},
\end{equation}
where $l_{i(\text{original})}$ denotes detected vanishing lines inside
$C$, $l_{i(\text{occluded})}$ denotes the recovered occluded boundary,
and $l_{i(\text{undetected})}$ denotes undetected vanishing lines
because of low contrast but recovered from geometric reasoning.  These
three types of lines are shown in Fig. \ref{fig:critical_line_floor}. 
With $l_{i(\text{original})}$ and vanishing point $v$, we generate all possible layouts $\textbf{L}$ using the model described
in Sec. \ref{overview}. 

\subsubsection{Layout Ranking}

\begin{figure}
\centering
\includegraphics[width=120mm]{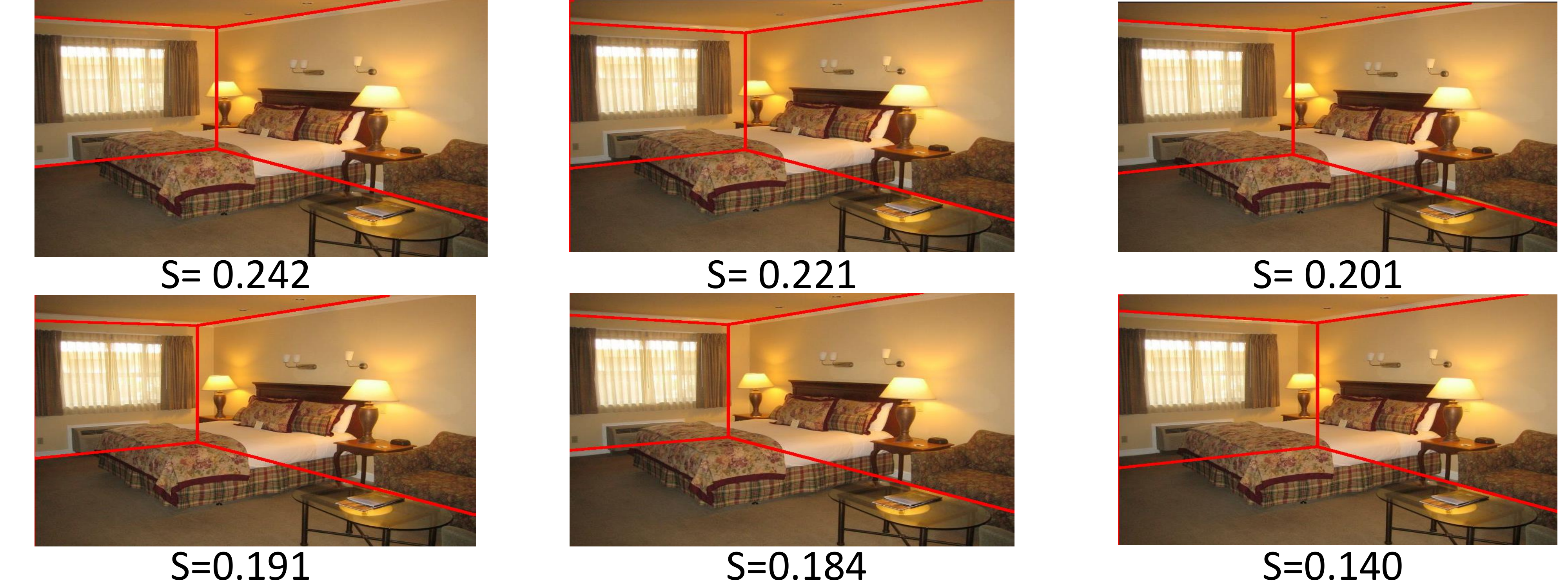} 
\caption{Example of Layout ranking using the proposed score function.}
\label{fig:layout_ranking}
\end{figure}

We use the coarse layout probability map $\textbf{P}$ as a weight mask to
evaluate the layout. The score function is defined as
\begin{equation}\label{equ:score}
S(\textbf{L}|\textbf{P}) = \frac{1}{N} \sum\limits_{i,j}\textbf{P}_{i,j}, 
\quad \forall \textbf{L}_{i,j}=1,
\end{equation}
where $\textbf{P}$ is the output from the MFCN, $\textbf{L}$ is a layout
from the hypotheses set, $N$ is a normalization factor that is equal to
the total number of layout pixels in $\textbf{L}$. Then, the optimal
layout is selected by
\begin{equation}\label{equ:opt}
\textbf{L}^{*} = \operatornamewithlimits{argmax}\limits_{\textbf{L}} 
S(\textbf{L}|\textbf{P}).
\end{equation}
The score function is in favor of the layout that is aligned well with
the coarse layout. Fig. \ref{fig:layout_ranking} shows one example where
the layout hypotheses are ranked using the score function in Eq.
(\ref{equ:opt}).  The layout with the highest score is chosen to be the
final result. 

\section{Experiments}\label{result}
 
\subsection{Experimental Setup}

We evaluate the proposed CFILE method on two popular datasets; namely,
Hedau's dataset \cite{hedau2009recovering} and the LSUN dataset
\cite{mallya2015learning}. Hedau dataset contains 209 training images, 53
validation images and 105 test images. Mallya {\em et al .} \cite{mallya2015learning} expanded Hedau dataset by adding 75 new images into training set while validation and test set unchanged, which referred to Hedau+ dataset. We conduct data augmentation for Hedau+ dataset as
done in \cite{mallya2015learning} by cropping, rotation, scaling and
luminance adjustment in the training of the MFCN. The LSUN dataset
\cite{mallya2015learning} contains 4000 training images, 394 validation
images and 1000 test images. Since no ground truth is released for the
1000 test images, we evaluate the proposed method on the 394 validation
set only. We resize all images to $404\times404$ by bicubic interpolation in the MFCN training,
and train two coarse layout models for the two datasets separately.

Hedau+ dataset provides both the layout and the geometric context
labels but it does not provide semantic surface labels. Thus, we use the
layout polygon provided in the dataset to generate semantic surface
labels. The LSUN dataset provides semantic surface labels but not the
layout. We detect edges on semantic surface labels and dilate them to a
width of 7 pixels in the MFCN training. By following
\cite{mallya2015learning}, we use the NYUDv2 RGBD dataset in 
\cite{gupta2013perceptual} for semantic segmentation to initialize the MFCN.
Also, we set the base learning rate to $10^{-4}$ with momentum $0.99$.

We adopt two performance metrics: the pixel-wise error and the corner
error. To compute the pixel-wise error, the obtained layout segmentation
is mapped to the ground truth layout segmentation. Then, the pixel-wise
error is the percentage of pixels that are wrongly matched. To compute
the corner error, we sum up all Euclidean distances between obtained
corners and their associated ground truth corners.  

\subsection{Experimental Results and Discussion}

The coarse layout scheme described in Sec. \ref{MFCN} is first
evaluated using the methodology in \cite{arbelaez2011contour}. We
compare our results, denoted by $\text{MFCN}_{1}$ and $\text{MFCN}_{2}$,
against the informative edge method \cite{mallya2015learning}, denoted
by FCN, in Table \ref{table:number_layout}. Our proposed two coarse
layout schemes have higher ODS (fixed contour threshold) and OIS
(per-image best threshold) scores. This indicates that they provide more
accurate regions for vanishing line samples in layout hypotheses
generation.  

\begin{table}[h]
\centering
\caption{Performance comparison of coarse layout results for Hedau's test
dataset, where the performance metrics are the fixed contour threshold
(ODS) and the per-image best threshold (OIS) \cite{arbelaez2011contour}.
We use $\text{FCN}$ to indicate the informative edge method in
\cite{mallya2015learning}. Both $\text{MFCN}_{1}$ and $\text{MFCN}_{2}$
are proposed in our work. They correspond to the two settings where the
layout and semantic surfaces are jointly trained on the original image
size ($\text{MFCN}_{1}$) and the downsampled image size $404\times404$.
($\text{MFCN}_{2}$)}\label{table:number_layout}
\begin{tabular}{|l|l|l|l|l|l|l|}
	\cline{2-7}
\multicolumn{1}{l|}{} & \multicolumn{2}{l|}{$\text{FCN}$\cite{mallya2015learning}}
&  \multicolumn{2}{l|}{$\text{MFCN}_{1}$(our)} 
& \multicolumn{2}{l|}{$\text{MFCN}_{2}$(our)} \\ \hline
Metrics & ODS   & OIS & ODS & OIS & ODS & OIS \\ \hline
Hedau's dataset  & 0.255 & 0.263 & 0.265 & 0.284& 0.265& 0.291\\ \hline
\end{tabular}
\end{table}

\begin{figure}[ht]
\centering
\includegraphics[width=120mm]{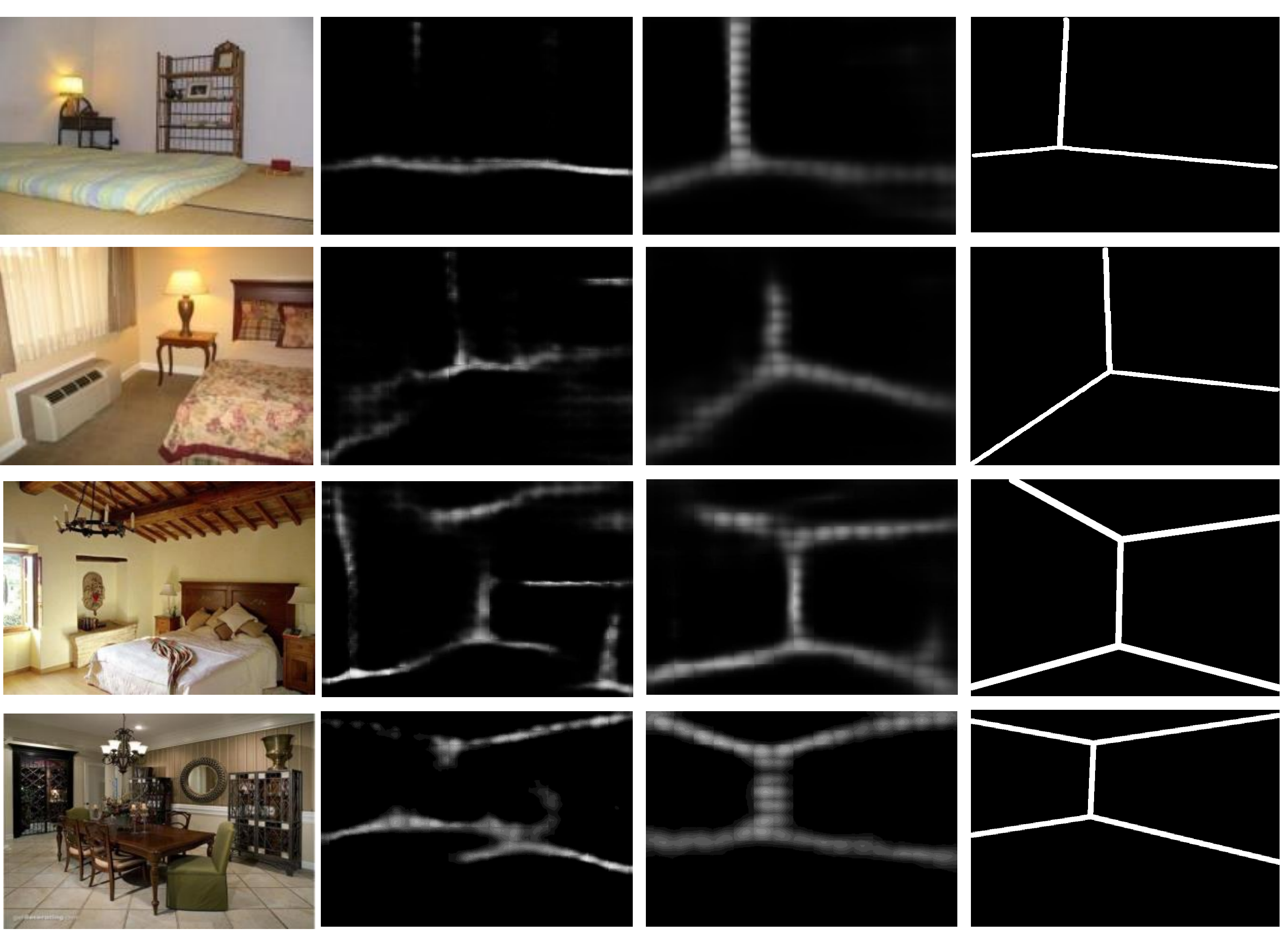} 
\caption{Comparison of coarse layout results (from left to right): the
input image, the coarse layout result of the FCN in \cite{mallya2015learning},
the coarse layout results of the proposed $\text{MFCN}_{2}$ and the ground
truth. The results of the $\text{MFCN}_{2}$ are more robust. Besides, it
provides clearer contours in occluded regions. The first two examples are from Hedau dataset and the last two examples are from LSUN dataset.} \label{fig:edge_result}
\end{figure}

We use several exemplary images to demonstrate that the proposed coarse
layout results are robust and close to the ground truth. That is, we
compare visual results of the FCN in \cite{mallya2015learning} and the
proposed $\text{MFCN}_{2}$ in Fig. \ref{fig:edge_result}. As compared to
the layout results of the FCN in \cite{mallya2015learning}, the proposed
$\text{MFCN}_{2}$ method provides robust and clearer layout results in
occluded regions, which are not much affected by object boundaries. 

\begin{figure}[htbp]
	\centering
	\includegraphics[width=120mm]{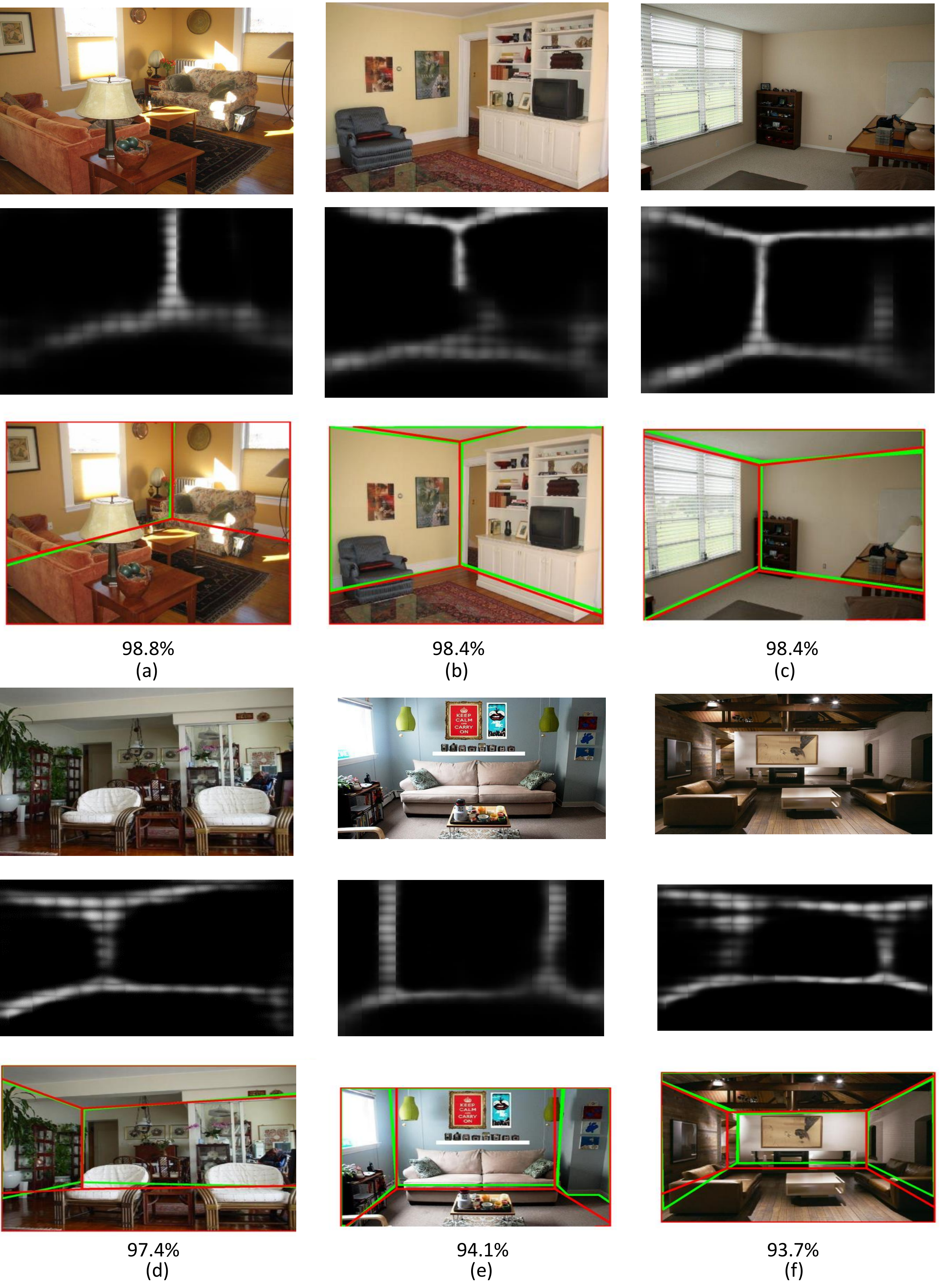} 
	\caption{Visualization of six best results of the CFILE method in
		Hedau's test dataset (from top to bottom): original images, the coarse
		layout estimates from MFCN, our results with pixel-wise accuracy (where the ground truth is shown in green
		and our result is shown in red).}\label{fig:result_visual_good}
\end{figure}

\begin{figure}[htbp]
	\centering
	\includegraphics[width=120mm]{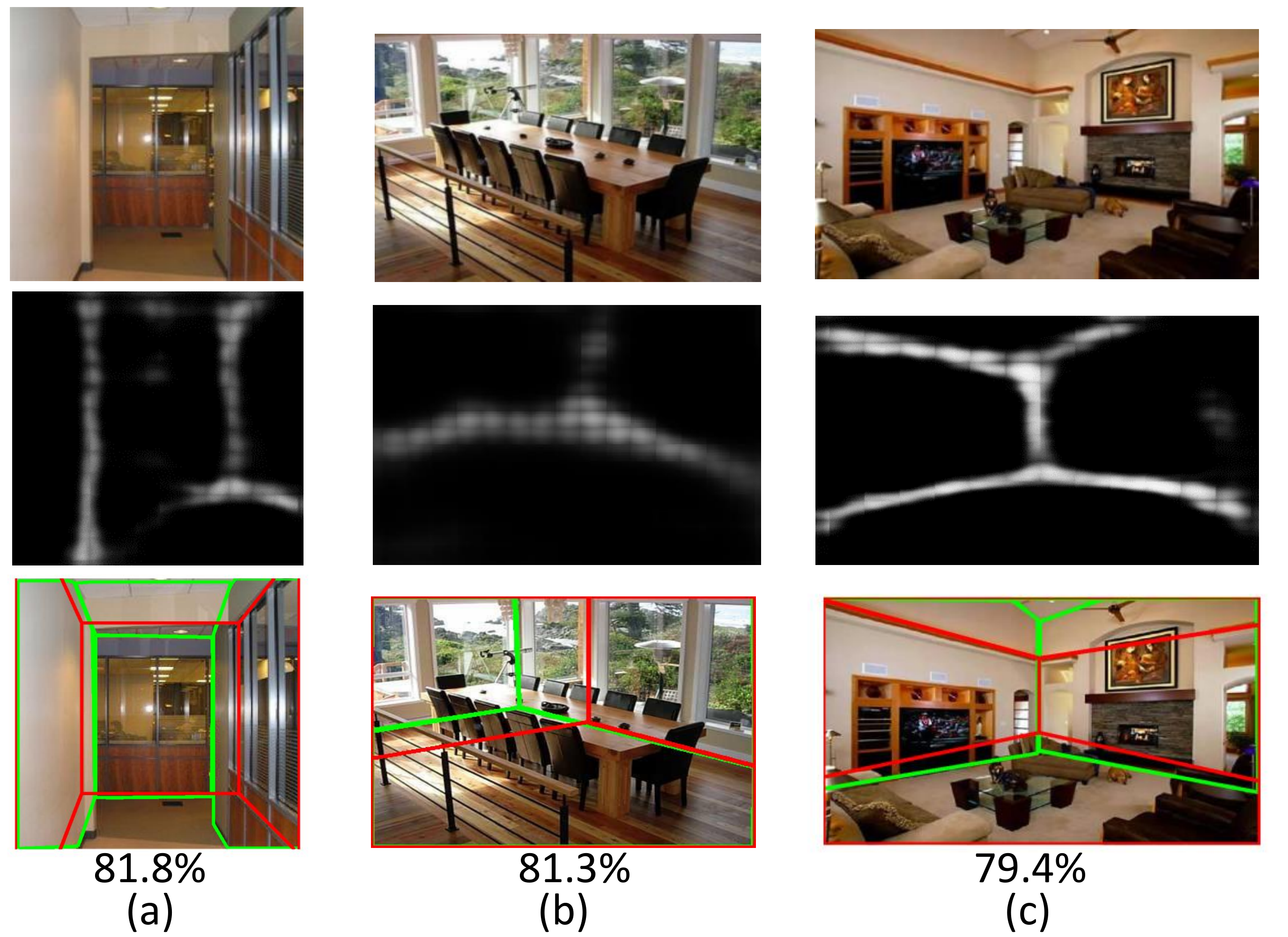} 
	\caption{Visualization of three worst results of the CFILE method in
		Hedau's test dataset (from top to bottom): original images, the coarse
		layout estimates from MFCN, our results with pixel-wise accuracy (where the ground truth is shown in green
		and our result is shown in red).}\label{fig:result_visual_bad}
\end{figure}

\begin{figure}[htbp]
	\centering
	\includegraphics[width=120mm]{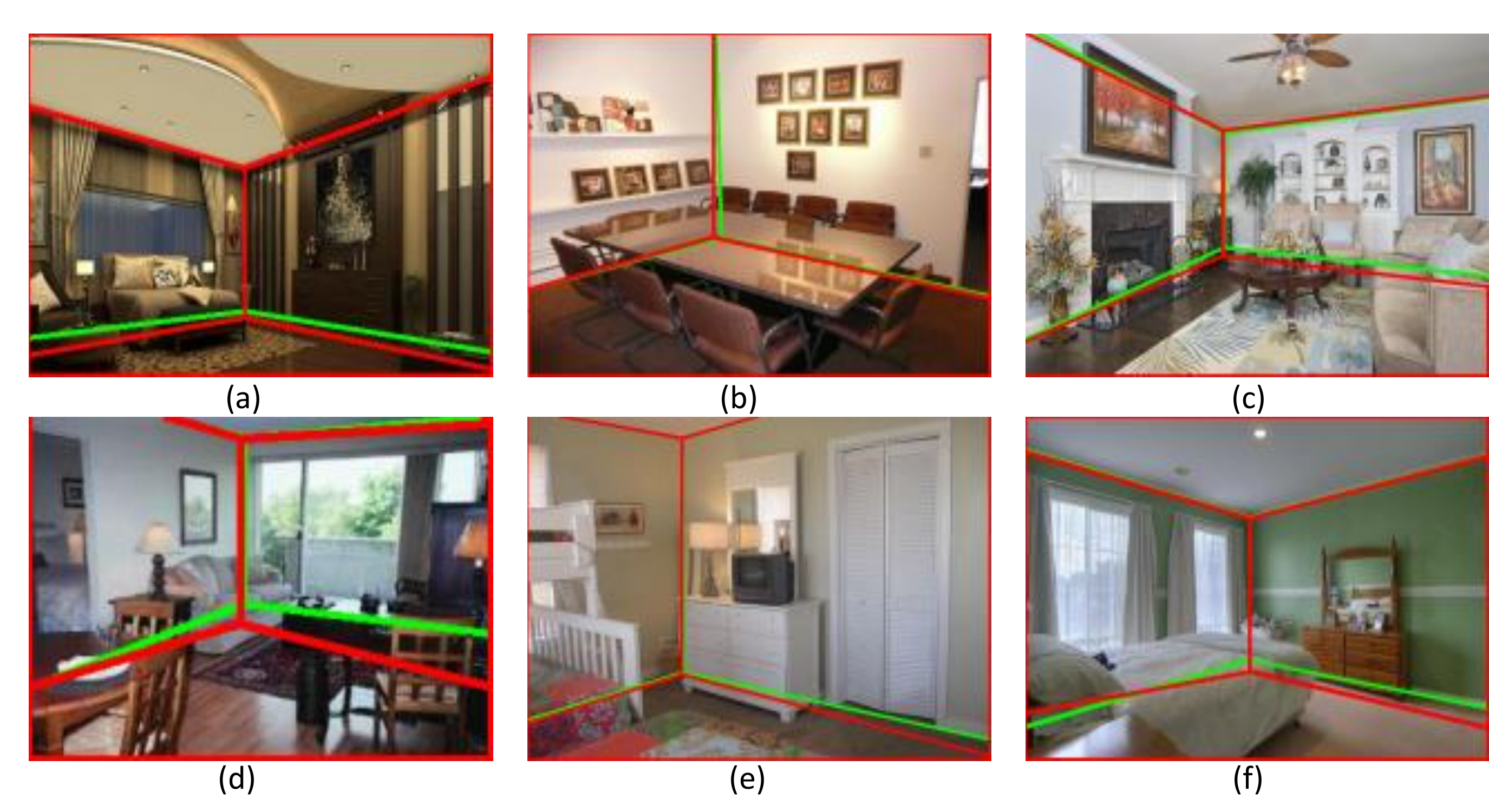} 
	\caption{Visualization of layout results of the CFILE method in
		the LSUN validation set. Ground truth is shown in green
		and our result is shown in red.}\label{fig:result_visual_LSUN}
\end{figure}

Next, we evaluate the performance of the proposed full layout algorithm,
CFILE, including the coarse layout estimation and the layout
optimization and ranking.  The performance of several methods for
Hedau's dataset and the LSUN dataset is compared in Table
\ref{table:number_hedau} and Table \ref{table:number_LSUN},
respectively.  The proposed CFILE method achieves state-of-the-art
performance. It outperforms the second best algorithm by $1.16\%$ in
Hedau's dataset and $1.32\%$ in the LSUN dataset.

\setlength{\tabcolsep}{4pt} 
\begin{table}
\begin{center}
\caption{Performance benchmarking for Hedau's dataset.}
\label{table:number_hedau}
\begin{tabular}{lr}\hline
{\bf Method} & {\bf Pixel Error (\%)}\\ \hline
Hedau {\em et al.} (2009)\cite{hedau2009recovering}  & 21.20\\
Del Pero {\em et al.} (2012)\cite{pero2012bayesian}  & 16.30\\
Gupta {\em et al.} (2010)\cite{gupta2010estimating}  & 16.20\\
Zhao {\em et al.} (2013)\cite{zhao2013scene}         & 14.50\\
Ramalingam {\em et al.} (2013)\cite{ramalingam2013manhattan}  & 13.34\\
Mallya {\em et al.} (2015)\cite{mallya2015learning}  & 12.83\\
Schwing {\em et al.} (2012)\cite{schwing2012efficient}  & 12.80\\
Del Pero {\em et al.} (2013)\cite{pero2013understanding}& 12.70\\
Dasgupta {\em et al.} (2016)\cite{Dasgupta2016Robust}   &     9.73\\
{\bf Proposed CFILE}                                    & {\bf 8.67} \\ \hline
\end{tabular}
\end{center}
\end{table}
\setlength{\tabcolsep}{1.4pt}

\setlength{\tabcolsep}{4pt}
\begin{table}
\begin{center}
\caption{Performance benchmarking for the LSUN dataset.}
\label{table:number_LSUN}
\begin{tabular}{lrr}\hline
{\bf Method} & {\bf Corner Error (\%)} & {\bf Pixel Error (\%)} \\ \hline
Hedau {\em et al.} (2009)\cite{hedau2009recovering} & 15.48 & 24.23\\
Mallya {\em et al.} (2015)\cite{mallya2015learning} & 11.02 & 16.71\\
Dasgupta {\em et al.} (2016) \cite{Dasgupta2016Robust}&8.20 & 10.63\\
{\bf Proposed CFILE} &  {\bf 7.95}& {\bf9.31} \\ \hline
\end{tabular}
\end{center}
\end{table}
\setlength{\tabcolsep}{1.4pt}

The best six results of the proposed CFILE method for Hedau's test images
are visualized in Fig. \ref{fig:result_visual_good}.  We see from these
five examples that the coarse layout estimation algorithm is robust in
highly cluttered rooms (see the second row and the fourth).  The layout refinement
algorithm can recover occluded boundaries accurately in Fig.
\ref{fig:result_visual_good} (a), (b), (d) and (e). It can also select
the best layout among several possible layouts. The worst three results of the proposed CFILE method for Hedau's test
images are visualized in Fig. \ref{fig:result_visual_bad}.  Fig.
\ref{fig:result_visual_bad} (a) show one example where the fine
layout result is misled by the wrong coarse layout estimate. Fig.
\ref{fig:result_visual_bad} (b) is a difficult case. The left wall and
right wall have the same appearance and there are several confusing wall
boundaries. Fig. \ref{fig:result_visual_bad} (c) gives the worst example of the CFILE method with accuracy $79.4\%$. However, it is still higher than the worst example reported in \cite{mallya2015learning} with accuracy
$61.05\%$. The ceiling boundary is confusing in Fig.
\ref{fig:result_visual_bad} (f). The proposed CFILE method selects the
ceiling line overlapping with the coarse layout. More visual results from the LSUN dataset are shown in Fig.
\ref{fig:result_visual_LSUN}.

\section{Conclusion and Future Work}\label{conclusion}

A coarse-to-fine indoor layout estimation (CFILE) method was proposed to
estimate the room layout from an RGB image. We adopted a multi-task
fully convolutional neural network (MFCN) to offer a robust coarse
layout estimate for a variety of indoor scenes with joint layout and
semantic surface training.  However, CNN is weak in enforcing spatial
constraints. To address this problem, we formulated an optimization
framework that enforces several constraints such as layout contour
straightness, surface smoothness and geometric constraints for layout
detail refinement. It was demonstrated by experimental results that the
proposed CFILE system offers the best performance on two commonly used
benchmark datasets.  It is an interesting topic to investigate the
multi-scale effect of CNN-based vision solutions and their applications
to semantic segmentation and geometrical layout of indoor scenes.

\bibliographystyle{splncs}
\bibliography{egbib}
 
\end{document}